\documentclass[conference]{IEEEtran}
\IEEEoverridecommandlockouts
\usepackage{cite}
\usepackage{amsmath,amssymb,amsfonts}
\usepackage{algorithmic}
\usepackage{graphicx}
\usepackage{textcomp}
\usepackage{xcolor}
\usepackage{tabularx} 

\usepackage{hyperref}       % hyperlinks
\usepackage{url}            % simple URL typesetting
\usepackage{booktabs}       % professional-quality tables
\usepackage{amsfonts}       % blackboard math symbols
\usepackage{nicefrac}       % compact symbols for 1/2, etc.
\usepackage{microtype}      % microtypography
\usepackage{xcolor}         % colors
\usepackage{amsmath,amssymb,amsfonts}
\usepackage{multirow}
\usepackage[table]{xcolor} % enables \cellcolor in tables

\usepackage{textcomp}
\usepackage{xcolor}
\usepackage{graphicx}
\usepackage{url}
\usepackage{subcaption} 
\usepackage{lipsum}  
\usepackage{booktabs}
\usepackage{hyperref}
\definecolor{BestGreen}{HTML}{C7E9C0}   % soft green (pick any you like)
\definecolor{SecondGreen}{HTML}{E5F5E0} % lighter green

\newcommand{\best}[1]{\cellcolor{BestGreen}\textbf{#1}}

\usepackage[table]{xcolor} % colors + safe in tables/captions
\DeclareRobustCommand{\tone}[1]{%
  \begingroup
  \setlength{\fboxsep}{0.6pt}%
  \colorbox{#1!25}{\strut\textcolor{black}{#1}}%
  \endgroup
}
\usepackage[ruled,vlined]{algorithm2e}
\usepackage{tikz}
\usetikzlibrary{arrows.meta,positioning,fit,calc,shapes.multipart}

\def\BibTeX{{\rm B\kern-.05em{\sc i\kern-.025em b}\kern-.08em
    T\kern-.1667em\lower.7ex\hbox{E}\kern-.125emX}}
\begin{document}

\title{LLM-Enhanced Reinforcement Learning for\\
Time Series Anomaly Detection
}

\author{\IEEEauthorblockN{1\textsuperscript{st} Bahareh Golchin}
\IEEEauthorblockA{\textit{dept. Computer Science} \\
\textit{Portland State University}\\
Portland, OR, USA \\
bgolchin@pdx.edu}
\and
\IEEEauthorblockN{2\textsuperscript{nd} Banafsheh Rekabdar}
\IEEEauthorblockA{\textit{dept. Computer Science} \\
\textit{Portland State University}\\
Portland, OR, USA  \\
rekabdar@pdx.edu}
\and
\IEEEauthorblockN{3\textsuperscript{rd} Danielle Justo}
\IEEEauthorblockA{\textit{dept. Computer Science} \\
\textit{Smith College}\\
Northampton, MA, USA \\
djusto@smith.edu
}

}

\maketitle

\begin{abstract}
Detecting anomalies in time series data is crucial for finance, healthcare, sensor networks, and industrial monitoring applications. However, time series anomaly detection often suffers from sparse labels, complex temporal patterns, and costly expert annotation. We propose a unified framework that integrates Large Language Model (LLM)–based potential functions for reward shaping with reinforcement learning (RL), Variational Autoencoder (VAE)–enhanced dynamic reward scaling, and active learning with label propagation. An LSTM-based RL agent leverages LLM-derived semantic rewards to guide exploration, while VAE reconstruction errors add unsupervised anomaly signals. Active learning selects the most uncertain samples, and label propagation expands labeled data efficiently. Evaluations on Yahoo-A1 and SMD benchmarks demonstrate that our method achieves state-of-the-art detection accuracy under limited labeling budgets and operates effectively in data-constrained settings. This study highlights the promise of combining LLMs with RL and advanced unsupervised techniques for robust, scalable anomaly detection in real-world applications.\footnote{Code is available at
  \href{https://github.com/baharehgl/Reward-Shaping-LLM/tree/main}{GitHub Repository}.}
\end{abstract}

\section{Introduction}
\label{sec1:Introduction}

Time series anomaly detection represents a critical challenge in modern data analysis, spanning applications from healthcare monitoring to industrial surveillance \cite{Kim2023ICSADSensors}. Traditional approaches face significant limitations when dealing with sparse labeled data, complex temporal dependencies, and real-time decision-making requirements \cite{Zhang2023SparseTS}. While deep learning methods show promise, they struggle with sample efficiency and adapting to novel anomaly patterns without extensive retraining. The convergence of reinforcement learning (RL) and anomaly detection represents a paradigm shift, treating anomaly detection as an interactive learning process where agents learn optimal detection policies through environmental feedback \cite{Bastani2025Tips}.

Existing RL-based anomaly detection systems face critical challenges, including sparse reward signals, limited exploration efficiency, and difficulty incorporating domain knowledge \cite{Kweider2024, Chen2024}. The scarcity and high cost of labeled anomaly data compound these limitations, as traditional RL methods require substantial labeled data to achieve satisfactory performance. These challenges require innovative approaches to operate effectively under data-constrained conditions while maintaining high detection accuracy \cite{Khanizadeh2025Smart}.

The emergence of Large Language Models (LLMs) has revolutionized complex reasoning tasks across diverse domains, offering new possibilities for enhancing RL-based anomaly detection systems. Recent research demonstrates that LLMs possess remarkable capabilities in understanding temporal patterns and can effectively serve as semantic anomaly detectors by leveraging their pre-trained knowledge to assess data patterns \cite{Bhambri2024Heuristics}. 

Potential-based reward shaping (PBRS) provides a theoretically sound framework for incorporating additional reward signals without altering optimal policies \cite{Ng1999}. This can prove to be particularly valuable in sparse reward environments where traditional RL algorithms struggle \cite{Forbes2024PBIM}. Active learning maximizes labeled sample informativeness through strategic selection, while techniques like label propagation expand labeled datasets. Variational Autoencoders (VAE) complement these approaches by providing unsupervised anomaly scoring through reconstruction error analysis \cite{Golchin2024AIxSET, Wiewiora2003}.

This work presents a novel integrated framework combining: an LSTM-based RL agent, VAE-enhanced reward mechanisms, LLM-based potential functions for semantic reward shaping, and active learning with label propagation. Our contributions include: (1) LLM-generated semantic potential functions incorporating domain knowledge without manual engineering, (2) dynamic reward scaling balancing supervised and unsupervised signals, and (3) comprehensive evaluation demonstrating superior performance on both univariate and multivariate datasets.

This comprehensive approach simultaneously addresses the sparse reward problem through LLM-based potential functions, sample efficiency through active learning, temporal dependencies through LSTM representation, and exploration-exploitation trade-offs through dynamic reward scaling. The resulting framework offers a robust solution for real-world anomaly detection where labeled data is scarce and timely, and accurate decision-making is critical.

\section{Background}
\label{sec3:background}

In what follows, we review the background related to our study.

\subsection{Extrinsic and Intrinsic Rewards}
% In RL, rewards are commonly divided into extrinsic and intrinsic categories. Extrinsic rewards are explicitly defined by the task designer and are given directly by the environment to drive the agent toward its main objective \cite{SuttonBarto2018, Ng1999}. For example, in anomaly detection, the agent may receive a positive extrinsic reward for correctly identifying an anomaly and a negative reward for false alarms or missed detections. These signals provide clear, task-specific feedback but can be sparse or delayed, making effective exploration more challenging \cite{Ng1999}.

% Intrinsic rewards, in contrast, are internally generated signals that encourage the agent to explore and learn even when extrinsic rewards are limited or absent \cite{Chentanez2005, Barto2013}. Such signals may be based on novelty, curiosity, or prediction error, motivating the agent to visit unfamiliar or informative states. In anomaly detection, intrinsic rewards can guide the agent toward recognizing unusual patterns that may not yet be labeled. Combining extrinsic and intrinsic rewards enables a balance between exploiting known strategies and exploring new situations, improving both learning efficiency and adaptability \cite{Pathak2017}.

In RL, rewards are categorized into extrinsic signals defined by the task designer and intrinsic signals generated internally to foster exploration \cite{SuttonBarto2018, Chentanez2005}. Extrinsic rewards drive the agent toward its primary objective, such as receiving positive feedback for correctly identifying anomalies or penalties for false alarms \cite{Ng1999}. Although these signals provide specific feedback, they are often sparse or delayed, which can hinder effective learning. Conversely, intrinsic rewards motivate the agent to investigate unfamiliar states based on factors like novelty or prediction error, independent of external feedback \cite{Barto2013}. For anomaly detection tasks, these intrinsic signals are crucial for guiding the agent toward unusual patterns that may not yet be labeled. By integrating both reward types, the agent effectively balances exploiting known strategies with exploring new situations, significantly enhancing both learning efficiency and adaptability \cite{Pathak2017}.

\subsection{Potential-Based Reward Shaping}

Reward shaping accelerates learning by modifying the reward signal while strictly preserving the optimal policy. A prominent approach, Potential-Based Reward Shaping (PBRS), augments the standard reward using a potential function $\Phi(s)$ defined as:
\begin{equation}
  r'(s,a,s') = r(s,a,s') + \gamma \Phi(s') - \Phi(s) 
  \label{eq:pbrs}
\end{equation}
\cite{Ng1999}. This formulation guarantees policy invariance, ensuring the agent's optimal behavior remains consistent with the original objective \cite{Wiewiora2003}. By incorporating domain heuristics or data-driven insights into $\Phi(s)$, PBRS guides the agent toward desirable states, such as identifying severe anomalies, more efficiently \cite{Devlin2012}. Consequently, the method improves convergence rates in complex environments without biasing the final learned policy \cite{Grzes2017}.

% VAEs map the original feature space to latent Gaussian distributions through an encoder-decoder architecture based on neural networks \cite{Elaziz2023DRLBusiness}. The learning objective is to maximize the intractable marginal likelihood $p(x;\theta)$, where \(x\) is a feature vector and \(\theta\) are decoder parameters. This is done by maximizing the Evidence Lower Bound (ELBO), which satisfies \(\log p_\theta(x) \ge \mathcal{L}(\theta,\phi; x)\):
% \begin{equation}
% \mathcal{L}(\theta,\phi; x)
% = \mathbb{E}_{q_\phi(z \mid x)} \big[ \log p_\theta(x \mid z) \big]
% - \mathrm{KL}\!\left( q_\phi(z \mid x) \,\|\, p(z) \right),
% \end{equation}
% where \(q_\phi(z \mid x)\) is the encoder (approximate posterior) and \(\mathrm{KL}[\cdot\|\cdot]\) denotes the Kullback–Leibler divergence between the approximate posterior and the prior \(p(z)\).

\subsection{Variational Autoencoder for Anomaly Detection}
VAEs map input features to latent Gaussian distributions using a neural encoder-decoder architecture \cite{Elaziz2023DRLBusiness}. The model maximizes the Evidence Lower Bound (ELBO) as a tractable proxy for the intractable marginal likelihood:
\begin{equation}
\mathcal{L}(\theta,\phi; x)
= \mathbb{E}_{q_\phi(z \mid x)} \big[ \log p_\theta(x \mid z) \big]
- \mathrm{KL}\!\left( q_\phi(z \mid x) \,\|\, p(z) \right),
\end{equation}
where the objective balances reconstruction fidelity with the Kullback–Leibler divergence between the approximate posterior \(q_\phi(z \mid x)\) and the prior \(p(z)\).

\section{Proposed Method}
\label{sec4:ProposedMethod}

\begin{figure*}[!t]
  \centering
  \includegraphics[width=\textwidth,height=0.22\textheight,keepaspectratio]{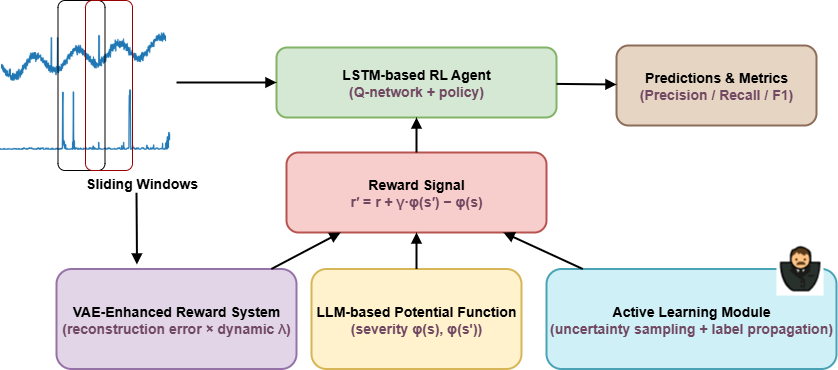}
  \caption{Our proposed framework: an LSTM-based RL agent operates on sliding windows of the time series; the reward merges a VAE-based reconstruction term and an LLM potential for semantic shaping with labels supplied via active learning, producing anomaly predictions.}
  \label{fig:proposed}
\end{figure*}

Our proposed framework illustrates in Figure \ref{fig:proposed}. The system consists of four main components: (1) an LSTM-based RL agent for sequential anomaly detection, (2) a VAE to estimate anomaly magnitude via reconstruction error with dynamic reward scaling, (3) an LLM-based semantic potential function for reward shaping, and (4) an active learning and label propagation module to minimize labeling cost. 

% \begin{figure}[htbp]
%     \centering
%     \includegraphics[width=0.80\textwidth]{Proposed-LLM.png}
%     \caption{Our proposed framework: an LSTM-based RL agent operates on sliding windows of the time series; the reward merges a VAE-based reconstruction term and an LLM potential for semantic shaping with labels supplied via active learning, producing anomaly predictions.}
%     \label{fig:proposed}
% \end{figure}

\subsection{LSTM-based RL agent }
\textbf{Environment and States.} Our framework models anomaly detection as a sequential decision-making process within an RL environment. The environment processes time series data sequentially, where at each time step $t$, the agent observes a state $s_t$, selects a binary action $a_t \in \{0, 1\}$ (normal or anomalous), and receives a reward signal. Each episode corresponds to processing a complete time series, with the agent's objective being to classify each time step while minimizing false detections accurately.

The state at time $t$ consists of a sliding window of $n_{\text{steps}}$ consecutive observations from the time series, $s_t = \{ x_{t-n_{\text{steps}}+1}, x_{t-n_{\text{steps}}+2}, \ldots, x_t \}$. This enables the agent to distinguish between different action contexts; each state is augmented with an action indicator flag. This creates two possible state representations at each time step: $s_t^0 = [s_t; 0]$ for predicting normal, and $s_t^1 = [s_t; 1]$ for predicting anomalous.
where each augmented state $s_t^a \in \mathbb{R}^{n_{\text{steps}} \times (d+1)}$ contains both the raw sensor values and the action flag. 

This dual-component design preserves essential temporal dependencies while providing the Q-network with action-specific context for learning distinct value functions. The sliding window length is set to $n_{\text{steps}} = 25$, providing sufficient historical context for temporal pattern recognition while maintaining computational efficiency.

\textbf{Extrinsic Reward.} The extrinsic reward mechanism provides direct feedback on the agent's classification decisions, guiding learning through immediate reinforcement based on prediction accuracy. At each time step, the agent selects a binary action $a_t \in \{0, 1\}$ representing predict normal or predict anomaly, respectively, and receives reward feedback based on the ground truth label $y_t$.

\textbf{Classification-Based Reward Structure.} The primary extrinsic reward $R_1(s_t, a_t)$ follows a confusion matrix-based formulation, giving $+5$ for true positives, $+1$ for true negatives, $-1$ for false positives, and $-5$ for false negatives ($R_1(s_t, a_t) \in \{5, 1, -1, -5\}$).

% that captures the four possible classification outcomes:
% \begin{equation}
% R_1(s_t, a_t) =
% \begin{cases}
% TP_{\text{val}} = 5, & \text{if } a_t = 1 \text{ and } y_t = 1 , \\
% TN_{\text{val}} = 1, & \text{if } a_t = 0 \text{ and } y_t = 0  \\
% FP_{\text{val}} = -1, & \text{if } a_t = 1 \text{ and } y_t = 0 \\
% FN_{\text{val}} = -5, & \text{if } a_t = 0 \text{ and } y_t = 1.
% \end{cases}
% \label{eq:extrinsic-reward}
% \end{equation}

% This asymmetric reward structure reflects the critical importance of anomaly detection in industrial applications, where missing an anomaly (false negative) typically incurs higher costs than raising a false alarm (false positive). The $5{:}1$ ratio between true positive and true negative rewards, along with the $5{:}1$ penalty ratio between false negatives and false positives, ensures the agent prioritizes anomaly detection sensitivity while maintaining reasonable specificity.

\subsection{VAE-Enhanced Reward Augmentation}
To complement the classification feedback, we incorporate reconstruction error from a VAE trained exclusively on normal patterns. The VAE provides an unsupervised anomaly score that captures deviations from learned normal behavior:
\begin{equation}
R_2(s_t) = \mathrm{MSE}(x_t, \hat{x}_t) = \frac{1}{n} \sum_{i=1}^{n} (x_{t,i} - \hat{x}_{t,i})^2,
\label{eq:vae-error}
\end{equation}
where $x_t$ represents the current window, $\hat{x}_t$ is its VAE reconstruction, and $n$ is the window dimensionality. Higher reconstruction errors indicate greater deviation from normal patterns, providing additional evidence for anomaly presence.

\textbf{Dynamic Reward Integration.} The final extrinsic reward combines classification accuracy with reconstruction-based guidance through a dynamically scaled formulation:
\begin{equation}
R_{\text{total}}(s_t, a_t) = R_1(s_t, a_t) + \lambda(t) \cdot R_2(s_t),
\label{eq:total-reward}
\end{equation}
where $\lambda(t)$ is a time-varying coefficient that balances supervised classification feedback ($R_1$) with unsupervised anomaly scoring ($R_2$). The coefficient $\lambda(t)$ is updated after each episode using proportional control:
\begin{equation}
\lambda_{t+1} = \mathrm{clip} \left( \lambda_t + \alpha \left( R_{\mathrm{target}} - R_{\mathrm{episode}} \right), \lambda_{\mathrm{min}}, \lambda_{\mathrm{max}} \right),
\label{eq:lambda-update}
\end{equation}
where $\alpha$ is the adjustment rate, $R_{\mathrm{target}}$ is the desired episode reward, and $R_{\mathrm{episode}}$ is the actual episode reward. If performance drops ($R_{\mathrm{episode}} < R_{\mathrm{target}}$), $\lambda$ increases to give more weight to the VAE anomaly signal; if performance improves, $\lambda$ decreases to focus more on classification. Bounds $[\lambda_{\mathrm{min}}, \lambda_{\mathrm{max}}]$ keep updates stable.

\subsection{LLM-Based Potential Function}

To enrich the reward signal with semantic context, we introduce a potential-based shaping term derived from an LLM. This component leverages the LLM's ability to assess complex temporal patterns and provide an interpretable anomaly severity score, which serves as the potential function $\phi(s)$ in the reward shaping framework as shown in Equation~\eqref{eq:pbrs}.

\textbf{Severity Scoring via LLM.} For a given state $s_t$, represented as a sliding window of $n_{\text{steps}}$ consecutive sensor readings, the values are formatted into a structured prompt and provided to the LLM (e.g., models GPT-3.5, Llama-3.2-3B, and Phi-2). The model is instructed to output a single JSON object: \{\text{"severity"}: v\}. Where $v \in [0,1]$ quantifies the likelihood of the current window containing an anomaly, with $0$ representing ``certainly normal'' and $1$ representing ``certainly anomalous''. This output is parsed to extract the scalar potential $\phi(s_t) = v$.

\textbf{Few-Shot Prompt Design.} The LLM prompt shows a few example sensor readings with their severity scores to help it give consistent and accurate results. For instance:
\begin{itemize}
    \item \texttt{Sensor readings: [0.0, 0.0, 0.0, 0.0, ...]} $\rightarrow$ \{\texttt{"severity": 0.00}\}
    \item \texttt{Sensor readings: [0.0, 0.0, 0.0, 5.0, 5.0, ...]} $\rightarrow$ \{\texttt{"severity": 0.75}\}
\end{itemize}
These examples act as anchors, which help the LLM to map diverse patterns to a bounded and meaningful anomaly severity scale. The prompt structure includes system-level instructions enforcing JSON-only output to prevent parsing errors and ensure reliable extraction of severity scores.

% \textbf{Caching and Precision Control.} To reduce computational overhead and ensure reproducibility, we apply two critical optimizations:
% \begin{enumerate}
%     \item \textbf{Rounding:} Each sensor value in $s_t$ is rounded to a fixed number of decimal places (default: 2) before being passed to the LLM. This mitigates minor numerical differences that would otherwise bypass caching, while maintaining sufficient precision for meaningful anomaly assessment.
%     \item \textbf{Caching:} The function $\phi(s_t)$ is memoized using an LRU cache (maximum 100{,}000 entries) so that identical rounded windows retrieve previously computed severity scores, avoiding repeated LLM queries. This dramatically reduces API costs and inference latency during training.
% \end{enumerate}

\textbf{Integration into Reward Shaping.} Recall Equation~\eqref{eq:pbrs} (i.e., $r'(s_t, a_t) = r(s_t, a_t) + \gamma \, \phi(s_{t+1}) - \phi(s_t)$). In our proposed method, we define: 1) $r(s_t, a_t)$ as the extrinsic reward from classification and VAE-based penalties, and 2) $\phi(s)$ as the LLM-generated potential. Next, similar to the literature, we define $\gamma$ as the discount factor. This approach preserves the optimal policy while providing a richer training signal that guides the agent toward states with higher anomaly likelihood.

\textbf{Semantic Context Enhancement.} Unlike traditional methods that only look for statistical patterns in data, the LLM potential function uses advanced reasoning to understand time-based patterns. The LLM can identify complex, unusual behaviors like sudden jumps, irregular changes, or subtle shifts in trends that purely numerical methods might miss. By combining basic numerical features with the LLM's high-level understanding, the potential function connects statistical anomaly detection with contextual reasoning. This helps the RL agent work more efficiently by better exploring and using informative areas of the data space. 
% Our LLM potential and shaped reward algorithm is detailed in Algorithm \ref{alg:LLM-reward}.

% \begin{algorithm}[t]
% \caption{LLM Potential $\phi(s)$ and Shaped Reward}
% \label{alg:LLM-reward}
% \KwIn{window $s \in \mathbb{R}^{n}$, next window $s'$, raw reward $r$, discount $\gamma$}
% \KwOut{shaped reward $r' = r + \gamma \phi(s') - \phi(s)$}

% \nl \textbf{function} \textsc{ComputePotential}$(s)$:\\
% \Indp
%     \nl Normalize $s$ to $\tilde{s}$ by z-scoring (fallback: all zeros if $\mathrm{std}(s)\approx0$)\;
%     \nl Build prompt with short guidelines and few-shot pairs; ask for JSON \verb|{"severity": s}|\;
%     \nl \textbf{if} LLM is GPT \textbf{then} call chat API; \textbf{else} call local HF pipeline\;
%     \nl Parse last numeric value in output; clamp to $[0,1]$; \textbf{if} parse fails \textbf{then} use small heuristic (e.g., last-point z-score)\;
%     \nl \textbf{return} $\phi(s)$\;
% \Indm
% \nl \textbf{function} \textsc{ShapedReward}$(r, s, s', \gamma)$:\\
% \Indp
%     \nl $\phi \leftarrow \textsc{ComputePotential}(s)$,\quad $\phi' \leftarrow \textsc{ComputePotential}(s')$\;
%     \nl \textbf{return} $r' \leftarrow r + \gamma \phi' - \phi$\;
% \Indm
% \end{algorithm}

\subsection{Active Learning Integration}

To reduce annotation effort while maintaining strong detection performance, the framework integrates an AL module that works alongside the RL agent. This module selects the most informative samples for human labeling and uses semi-supervised learning to spread labels to similar unlabeled instances. At the end of each episode, uncertainty is measured by the margin $\mathrm{Margin}(s) = \left| Q(s, a_0) - Q(s, a_1) \right|$,
% \begin{equation}
% \mathrm{Margin}(s) = \left| Q(s, a_0) - Q(s, a_1) \right|,
% \label{eq:margin}
% \end{equation}
and the top $N_{\mathrm{AL}}$ states with the smallest margins are sent for manual annotation. A label propagation step then assigns pseudo-labels to additional unlabeled samples based on similarity weights $w_{ij}$, typically computed with a Gaussian kernel in feature space. The $K_{\mathrm{LP}}$ most confident pseudo-labels are added to the labeled set, further improving the training signal while limiting human labeling effort.

\section{Experiments}
\label{sec5:Experiment}

This section describes our experiments. We start by describing the datasets and then address three key research questions: (\textbf{Q1}) Does using LLMs for potential-based reward shaping perform better than existing methods? (\textbf{Q2}) Which LLM (GPT 3.5, Llama-3, Phi-2) works best in our anomaly detection approach and why? (\textbf{Q3}) How does our method perform on univariate versus multivariate time series datasets?

% This section outlines our experiments. We first begin with dataset specifications, and then aim to answer three main research questions: (\textbf{Q1}) Can using LLMs as potential-based reward shaping improve upon the existing methods in the literature? (\textbf{Q2}) Which LLM (GPT 3.5, Llama-3, Phi-2) performs better in our proposed anomaly detection framework? And why? (\textbf{Q3}) Is our proposed method sensitive to univariate or multivariate time series datasets?

\subsection{Datasets}

We evaluate our framework on two widely used benchmarks: Yahoo-A1 (67 univariate series from real Yahoo website traffic) and SMD (28 multivariate series with 38 sensors from server machines). Yahoo-A1 contains hourly collected data with labeled anomaly points, while SMD includes 10 days of sensor data with normal patterns in the first 5 days and random anomalies in the last 5 days \cite{Laptev2015Yahoo, Su2019SRNN}. Table \ref{tab:yahoo-smd-stats} summarizes the key dataset statistics.

\begin{table}[htbpt]
  \centering
  \small
  \caption{Key Statistics for Yahoo-A1 and SMD}
  \label{tab:yahoo-smd-stats}
  \begin{tabular}{lrrr}
    \toprule
    \textbf{Benchmark} & \textbf{\# series} & \textbf{\# dims} & \textbf{Anomaly \%} \\
    \midrule
    Yahoo-A1  & 67  & 1  & 1.76\% \\
    SMD  & 28  & 38  & 4.16\% \\
    \bottomrule
  \end{tabular}
\end{table}

\subsection{Results and Discussions}
To answer \textbf{Q1}, Table~\ref{tab:results} (left) summarizes performance on Yahoo A1. Our method with Llama-3 achieves an F1 of 0.7413 with precision/recall of 0.6051/0.9565, essentially on par with the strongest non-LLM baseline (CARLA: 0.7233 F1). Compared with earlier deep baselines such as TimesNet (0.5135 F1) and TranAD (0.5654 F1), the gain is sizeable. The high recall indicates the LLM-shaped potential guides the agent to consistently surface true anomalies, while maintaining moderate precision. In contrast, the GPT-3.5 variant attains very high recall (0.9130) but extremely low precision (0.0742), leading to a poor F1 of 0.1372. This reflects an over-eager anomaly prior to GPT that over-flags peaks. The Phi-2 variant shows the opposite behavior, 0.6666 precision but only 0.4761 recall, suggesting a conservative prior that misses many positives; its F1 is 0.5555.

On SMD (right side of the Table~\ref{tab:results}), Llama-3 again provides the best trade-off, reaching an F1 of 0.5300 with precision/recall of 0.3813/0.8685. This outperforms CARLA (0.5114 F1) and other non-LLM baselines. The Industrial SMD series contains strong seasonality and gradual drifts; the Llama-shaped potential appears calibrated enough to keep recall high without flooding detections. GPT-3.5 performs better than on Yahoo (precision/recall 0.5370/0.4061; F1 = 0.4625), but it still underperforms Llama-3, likely because its anomaly prior is too permissive. Phi-2 is highly precise on SMD (0.8461) yet recalls only 0.2541 of anomalies (F1 0.3908), again consistent with a cautious prior.

\begin{figure}[!t]
  \centering
  \includegraphics[width=\columnwidth,keepaspectratio,trim=0 2mm 0 1mm,clip]{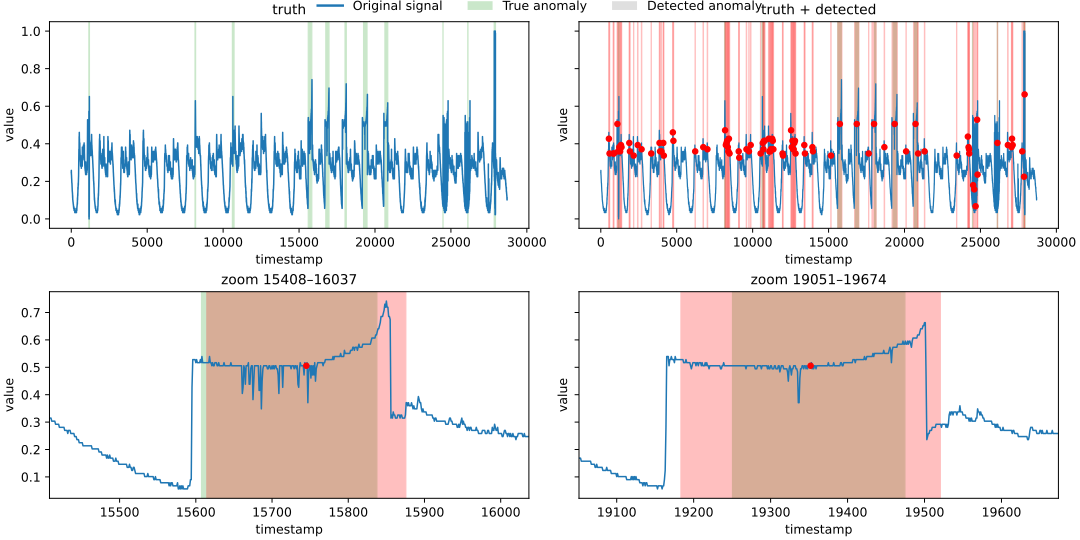}
  \caption{SMD anomaly detection — Llama. Colors: \tone{blue} = original signal; \tone{green} = true (ground-truth) anomalies; \tone{red} = detected anomalies. Panels: top-left shows ground truth over the signal; top-right shows ground truth + detected anomalies; bottom-left and bottom-right are two zoomed-in regions.}
  \label{fig:SMD-LLAMA}
\end{figure}

To answer \textbf{Q2}, we identify three practical reasons. First, Llama-3 follows instructions well and produces smooth severity scores within the required $[0,1]$ range, creating balanced signals that work effectively for potential-based reward shaping. Second, Llama-3 shows better pattern recognition compared to other models. GPT often overreacts to minor changes and assigns high scores too broadly. 
An example of this phenomenon is illustrated in 
% Figure \ref{fig:SMD-GPT}. On the contrary, Phi-2 underestimates moderate changes and misses important events as shown in the Figure \ref{fig:SMD-PHI}. 
Figure \ref{fig:SMD-LLAMA}, where Llama-3 properly identifies significant changes and outliers without being overly sensitive to normal variations. Third, Llama-3 produces more consistent outputs across different time windows, resulting in less noisy shaped rewards and helping the RL agent learn clearer decision boundaries.

% \begin{figure}[!t]
%   \centering
%   \includegraphics[width=\columnwidth,keepaspectratio,trim=0 2mm 0 1mm,clip]{SMD_GPT.png}
%   \caption{SMD anomaly detection — GPT. Colors: \tone{blue} = original signal; \tone{green} = true (ground-truth) anomalies; \tone{red} = detected anomalies. Panels: top-left shows ground truth over the signal; top-right shows ground truth + detected anomalies; bottom-left and bottom-right are two zoomed-in regions.}
%   \label{fig:SMD-GPT}
% \end{figure}

% \begin{figure}[!t]
%   \centering
%   \includegraphics[width=\columnwidth,keepaspectratio,trim=0 2mm 0 1mm,clip]{SMD_Phi.png}
%   \caption{SMD anomaly detection — Phi. Colors: \tone{blue} = original signal; \tone{green} = true (ground-truth) anomalies; \tone{red} = detected anomalies. Panels: top-left shows ground truth over the signal; top-right shows ground truth + detected anomalies; bottom-left and bottom-right are two zoomed-in regions.}
%   \label{fig:SMD-PHI}
% \end{figure}

% \begin{figure}[htbp]
%     \centering
%     \includegraphics[width=0.90\textwidth]{SMD_GPT.png}
%     \caption{Example visualizations of SMD anomaly detection results of GPT from the proposed method.}
%     \label{fig:SMD-GPT}
% \end{figure}

% \begin{figure}[htbp]
%     \centering
%     \includegraphics[width=0.90\textwidth]{SMD_Phi.png}
%     \caption{Example visualizations of SMD anomaly detection results of Phi from the proposed method.}
%     \label{fig:SMD-PHI}
% \end{figure}

% \begin{figure}[htbp]
%     \centering
%     \includegraphics[width=0.90\textwidth]{LLAMA.png}
%     \caption{Example visualizations of SMD anomaly detection results of Llama from the proposed method.}
%     \label{fig:SMD-LLAMA}
% \end{figure}

To answer \textbf{Q3}, Yahoo A1 is a simple dataset where each time series has only one variable and anomalies are typically sudden spikes or shifts in the mean value. These types of anomalies are easy to detect using short time windows because an LLM can identify obvious patterns like spikes or jumps, leading to good detection performance. In contrast, SMD is a complex multivariate dataset with many related sensors per machine that show seasonal patterns and gradual changes over time. In SMD, real anomalies usually occur when multiple sensors change together or when deviations continue for extended periods, making individual sensors appear normal despite the overall anomaly. This complexity makes detection more difficult because methods need to understand relationships between different sensors and track changes over time, or they will either miss real anomalies or incorrectly flag normal seasonal variations. Our approach computes LLM severity scores for each window and uses them for reward shaping, which works well for Yahoo's simple, single-variable anomalies but struggles with SMD's complex multi-sensor patterns unless the model can capture cross-sensor relationships and temporal consistency.

\begin{table*}[!t]
\caption{Comparison of our framework on two benchmark time series datasets: Yahoo A1 and SMD}
\label{tab:results}
\centering
\setlength{\tabcolsep}{5pt}            % horizontal padding
\renewcommand{\arraystretch}{1.15}     % row height
\footnotesize                          % or \scriptsize if you need it tighter

\begin{tabular*}{\textwidth}{@{\extracolsep{\fill}} lccc ccc}
\toprule
& \multicolumn{3}{c}{\textbf{Yahoo A1}} & \multicolumn{3}{c}{\textbf{SMD}} \\
\cmidrule(lr){2-4}\cmidrule(lr){5-7}
\textbf{Model} & \textbf{Prec} & \textbf{Rec} & \textbf{F1} &
\textbf{Prec} & \textbf{Rec} & \textbf{F1} \\
\midrule
THOC \cite{Shen2020THOC}                 & 0.1495 & 0.8326 & 0.2534 & 0.0997 & 0.5307 & 0.1679 \\
TranAD \cite{Tuli2022TranAD}             & 0.4185 & 0.8712 & 0.5654 & 0.2649 & 0.5661 & 0.3609 \\
TS2Vec \cite{Yue2022TS2Vec}              & 0.3929 & 0.6305 & 0.4841 & 0.1033 & 0.5295 & 0.1728 \\
DCdetector \cite{Yang2023DCdetector}     & 0.0598 & 0.9434 & 0.1124 & 0.0432 & 0.9967 & 0.0828 \\
TimesNet \cite{Wu2023TimesNet}           & 0.3808 & 0.7883 & 0.5135 & 0.2450 & 0.5474 & 0.3385 \\
CARLA \cite{Darban2025CSUR}              & 0.5747 & 0.9755 & 0.7233 & 0.4276 & 0.6362 & 0.5114 \\
\midrule
Proposed Method + GPT-3.5                & 0.0742 & 0.9130 & 0.1372 & 0.5370 & 0.4061 & 0.4625 \\
Proposed Method + Llama-3.2-3B           & 0.6051 & 0.9565 & \best{0.7413} & 0.3813 & 0.8685 & \best{0.5300} \\
Proposed Method + Phi-2                  & 0.6666 & 0.4761 & 0.5555 & 0.8461 & 0.2541 & 0.3908 \\
\bottomrule
\end{tabular*}
\end{table*}

% \begin{table}[htbp]
% \centering
% \caption{Comparison of our framework on two benchmark time series datasets: Yahoo A1 and SMD}
% \label{tab:results}
% \setlength{\tabcolsep}{4pt} % tighten columns for two-column format
% \small
% \begin{tabular}{lccc ccc}
% \toprule
% & \multicolumn{3}{c}{\textbf{Yahoo A1}} & \multicolumn{3}{c}{\textbf{SMD}} \\
% \cmidrule(lr){2-4}\cmidrule(lr){5-7}
% \textbf{Model} & \textbf{Prec} & \textbf{Rec} & \textbf{F1} &
% \textbf{Prec} & \textbf{Rec} & \textbf{F1} \\
% \midrule
% THOC (\cite{Shen2020THOC})                 & 0.1495 & 0.8326 & 0.2534 & 0.0997 & 0.5307 & 0.1679 \\
% TranAD (\cite{Tuli2022TranAD})               & 0.4185 & 0.8712 & 0.5654 & 0.2649 & 0.5661 & 0.3609\\
% TS2Vec (\cite{Yue2022TS2Vec})               & 0.3929 & 0.6305 & 0.4841 & 0.1033 &  0.5295 & 0.1728\\
% DCdetector (\cite{Yang2023DCdetector})           & 0.0598 & 0.9434 & 0.1124 & 0.0432 & 0.9967 & 0.0828\\
% TimesNet (\cite{Wu2023TimesNet})             & 0.3808 & 0.7883 & 0.5135 & 0.2450 & 0.5474 & 0.3385 \\
% CARLA (\cite{Darban2025CSUR})                & 0.5747 & 0.9755 & 0.7233 & 0.4276 & 0.6362 & 0.5114\\
% \midrule
% Proposed Method + GPT-3.5              & 0.0742 & 0.9130 & 0.1372 & 0.5370 & 0.4061 & 0.4625\\
% Proposed Method + Llama-3.2-3B          & 0.6051 & 0.9565 & \best{0.7413} & 0.3813 & 0.8685 & \best{0.5300}\\
% Proposed Method + Phi-2              & 0.6666 & 0.4761 & 0.5555 & 0.8461 & 0.2541 & 0.3908\\
% \bottomrule
% \end{tabular}
% \end{table}

\section{Conclusion}
\label{sec6:Conclusion}
This study presents a novel framework integrating LLM-based potential functions with reinforcement learning for time series anomaly detection, addressing sparse rewards and limited labeled data challenges. Our approach demonstrates that semantic reward shaping through LLMs, particularly Llama-3, effectively guides RL agents while maintaining policy invariance, with VAE-enhanced dynamic reward scaling and active learning achieving competitive performance on Yahoo-A1 and SMD datasets. Experimental results reveal distinct LLM behavioral patterns in anomaly assessment, with Llama-3 providing optimal precision-recall balance across univariate and multivariate scenarios. The framework's effectiveness under data-constrained conditions makes it valuable for real-world applications, with future work exploring complex multivariate relationships and additional LLM architectures.

\section*{Acknowledgment}

This material is based on work supported by the National Science Foundation under grant no. 2244551.

\bibliographystyle{IEEEtran}
\bibliography{refs}

@book{SuttonBarto2018,
  author    = {Richard S. Sutton and Andrew G. Barto},
  title     = {Reinforcement Learning: An Introduction},
  edition   = {2},
  publisher = {MIT Press},
  address   = {Cambridge, MA, USA},
  year      = {2018}
}

@inproceedings{Ng1999,
  author    = {Andrew Y. Ng and Daishi Harada and Stuart J. Russell},
  title     = {Policy Invariance under Reward Transformations: Theory and Application to Reward Shaping},
  booktitle = {Proc. 16th Int. Conf. Mach. Learn. (ICML)},
  pages     = {278--287},
  year      = {1999}
}

@incollection{Chentanez2005,
  author    = {N. Chentanez and Andrew G. Barto and Satinder P. Singh},
  title     = {Intrinsically Motivated Reinforcement Learning},
  booktitle = {Advances in Neural Information Processing Systems 17},
  publisher = {MIT Press},
  address   = {Cambridge, MA, USA},
  pages     = {1281--1288},
  year      = {2005}
}

@incollection{Barto2013,
  author    = {Andrew G. Barto},
  title     = {Intrinsic Motivation and Reinforcement Learning},
  booktitle = {Intrinsic Motivation and Reinforcement Learning},
  editor    = {G. Baldassarre and M. Mirolli},
  publisher = {Springer},
  address   = {Berlin, Germany},
  pages     = {17--47},
  year      = {2013}
}

@inproceedings{Pathak2017,
  author    = {Deepak Pathak and Pulkit Agrawal and Alexei A. Efros and Trevor Darrell},
  title     = {Curiosity-Driven Exploration by Self-Supervised Prediction},
  booktitle = {Proc. IEEE Conf. Comput. Vis. Pattern Recognit. (CVPR)},
  pages     = {2778--2787},
  year      = {2017}
}

@article{Wiewiora2003,
  author    = {Eric Wiewiora},
  title     = {Potential-Based Shaping and Q-Value Initialization Are Equivalent},
  journal   = {J. Artif. Intell. Res.},
  volume    = {19},
  pages     = {205--208},
  year      = {2003}
}

@inproceedings{Devlin2012,
  author    = {Sam Devlin and Daniel Kudenko},
  title     = {Dynamic Potential-Based Reward Shaping},
  booktitle = {Proc. 11th Int. Conf. Autonomous Agents and Multiagent Systems (AAMAS)},
  pages     = {433--440},
  year      = {2012}
}

@inproceedings{Grzes2017,
  author    = {Marek Grzes},
  title     = {Reward Shaping in Episodic Reinforcement Learning},
  booktitle = {Proc. 16th Int. Conf. Autonomous Agents and MultiAgent Systems (AAMAS)},
  pages     = {565--573},
  year      = {2017}
}

@online{Laptev2015Yahoo,
  author    = {Nikolay Laptev and Brian Y. and S. Amizadeh},
  title     = {A Benchmark Dataset for Time Series Anomaly Detection},
  year      = {2015},
  note      = {Yahoo! Research Blog},
  url       = {https://yahooresearch.tumblr.com/post/114590420346/a-benchmark-dataset-for-time-series-anomaly},
  urldate   = {2025-10-08}
}

@inproceedings{Su2019SRNN,
  author    = {Yujun Su and Yunzhi Zhao and Chaochao Niu and Rong Liu and Wei Sun and Dan Pei},
  title     = {Robust Anomaly Detection for Multivariate Time Series via Stochastic Recurrent Neural Network},
  booktitle = {Proc. ACM SIGKDD Int. Conf. Knowl. Discov. Data Min. (KDD)},
  pages     = {2828--2837},
  year      = {2019}
}

@article{Kweider2024,
  author    = {Louay Kweider and Mustafa Abou Kassem and Usama Sandouk},
  title     = {Anomalous State Sequence Modeling to Enhance Safety in Reinforcement Learning},
  journal   = {IEEE Access},
  volume    = {12},
  pages     = {157140--157148},
  year      = {2024},
  doi       = {10.1109/ACCESS.2024.3486549}
}

@misc{Chen2024,
  author    = {Xiang Chen and Rui Xiao and Zhiqiang Zeng and Zilong Qiu and Shukai Zhang and Xuefeng Du},
  title     = {Semi-supervised Anomaly Detection via Adaptive Reinforcement Learning-Enabled Method with Causal Inference for Sensor Signals},
  eprint    = {2405.06925},
  archivePrefix = {arXiv},
  primaryClass = {cs.LG},
  year      = {2024}
}

@misc{Bhambri2024Heuristics,
  author    = {Shreyas Bhambri and Aniket Bhattacharjee and Deepanshu Kalwar and Liang Guan and Hongxin Liu and Subbarao Kambhampati},
  title     = {Extracting Heuristics from Large Language Models for Reward Shaping in Reinforcement Learning},
  eprint    = {2405.15194},
  archivePrefix = {arXiv},
  primaryClass = {cs.LG},
  year      = {2024}
}

@inproceedings{Forbes2024PBIM,
  author    = {G. C. Forbes and N. Gupta and L. Villalobos-Arias and C. M. Potts and A. Jhala and D. L. Roberts},
  title     = {Potential-Based Reward Shaping for Intrinsic Motivation},
  booktitle = {Proc. 23rd Int. Conf. Autonomous Agents and Multiagent Systems (AAMAS)},
  pages     = {589--597},
  address   = {Auckland, New Zealand},
  year      = {2024}
}

@inproceedings{Golchin2024AIxSET,
  author    = {Bahareh Golchin and Banafsheh Rekabdar},
  title     = {Anomaly Detection in Time Series Data Using Reinforcement Learning, Variational Autoencoder, and Active Learning},
  booktitle = {Proc. Conf. AI, Science, Engineering, and Technology (AIxSET)},
  pages     = {1--8},
  address   = {Laguna Hills, CA, USA},
  year      = {2024},
  publisher = {IEEE}
}

@article{Kim2023ICSADSensors,
  author    = {Bumsoo Kim and Mohammed A. Alawami and Eunyoung Kim and Sungho Oh and Jihun Park and Hyoungseob Kim},
  title     = {A Comparative Study of Time Series Anomaly Detection Models for Industrial Control Systems},
  journal   = {Sensors},
  volume    = {23},
  number    = {3},
  pages     = {1310},
  year      = {2023}
}

@inproceedings{Zhang2023SparseTS,
  author    = {Mengying Zhang and Yifan Sun and Feng Liang},
  title     = {Sparse Deep Learning for Time Series Data: Theory and Applications},
  booktitle = {Proc. 37th Conf. Neural Inf. Process. Syst. (NeurIPS 2023), Poster},
  year      = {2023}
}

@article{Bastani2025Tips,
  author    = {Hamsa Bastani and Osbert Bastani and Warut S. Park Sinchaisri},
  title     = {Improving Human Sequential Decision-Making with Reinforcement Learning},
  journal   = {Management Science},
  note      = {Articles in Advance},
  year      = {2025}
}

@article{Khanizadeh2025Smart,
  author    = {Fatemeh Khanizadeh and Amirhossein Ettefaghian and Graham Wilson and Amir Shirazibeheshti and Tarek Radwan and Cristian Luca},
  title     = {Smart Data-Driven Medical Decisions Through Collective and Individual Anomaly Detection in Healthcare Time Series},
  journal   = {Int. J. Med. Informatics},
  volume    = {194},
  pages     = {105696},
  year      = {2025}
}

@article{Elaziz2023DRLBusiness,
  author    = {Eid A. Elaziz and Rania Fathalla and Mohamed Shaheen},
  title     = {Deep Reinforcement Learning for Data-Efficient Weakly Supervised Business Process Anomaly Detection},
  journal   = {Journal of Big Data},
  volume    = {10},
  number    = {1},
  pages     = {33},
  year      = {2023}
}

@article{Darban2025CSUR,
  author    = {Zahra Zamanzadeh Darban and Geoffrey I. Webb and Shirui Pan and Charu C. Aggarwal and Mahdieh Soleimani},
  title     = {Deep Learning for Time Series Anomaly Detection: A Survey},
  journal   = {ACM Comput. Surv.},
  volume    = {57},
  number    = {1},
  pages     = {Article 15, 1--42},
  year      = {2025}
}

@inproceedings{Shen2020THOC,
  author    = {Li Shen and Zhenyu Li and James T. Kwok},
  title     = {Time-Series Anomaly Detection Using Temporal Hierarchical One-Class Network},
  booktitle = {Advances in Neural Information Processing Systems 33 (NeurIPS)},
  pages     = {13016--13026},
  year      = {2020}
}

@article{Tuli2022TranAD,
  author    = {Shreshth Tuli and Giuliano Casale and Nicholas R. Jennings},
  title     = {TranAD: Deep Transformer Networks for Anomaly Detection in Multivariate Time Series Data},
  journal   = {Proc. VLDB Endow.},
  volume    = {15},
  number    = {6},
  pages     = {1201--1214},
  year      = {2022}
}

@inproceedings{Yue2022TS2Vec,
  author    = {Zhenhui Yue and Yuxuan Wang and Jiaqi Duan and Tao Yang and Chao Huang and Yuxiao Tong and Bo Xu},
  title     = {TS2Vec: Towards Universal Representation of Time Series},
  booktitle = {Proc. AAAI Conf. Artif. Intell. (AAAI)},
  pages     = {8980--8987},
  year      = {2022}
}

@inproceedings{Yang2023DCdetector,
  author    = {Yonggang Yang and Chulin Zhang and Tian Zhou and Qingsong Wen and Liang Sun},
  title     = {{DCdetector}: Dual Attention Contrastive Representation Learning for Time Series Anomaly Detection},
  booktitle = {Proc. 29th ACM SIGKDD Int. Conf. Knowl. Discov. Data Min. (KDD)},
  address   = {Long Beach, CA, USA},
  year      = {2023}
}

@inproceedings{Wu2023TimesNet,
  author    = {Haixu Wu and Tengfei Hu and Yuxuan Liu and Haoran Zhou and Jianmin Wang and Mingsheng Long},
  title     = {TimesNet: Temporal 2D-Variation Modeling for General Time Series Analysis},
  booktitle = {Int. Conf. Learn. Representations (ICLR)},
  year      = {2023}
}

\end{document}